\def\year{2022}\relax
\documentclass[letterpaper]{article} 
\usepackage{aaai22}  
\usepackage{times}  
\usepackage{helvet}  
\usepackage{courier}  
\usepackage[hyphens]{url}  
\usepackage{graphicx} 
\usepackage{natbib}  
\usepackage{caption} 
\DeclareCaptionStyle{ruled}{labelfont=normalfont,labelsep=colon,strut=off} 
\usepackage{algorithm}
\usepackage{algorithmic}

\usepackage{newfloat}
\usepackage{listings}
\floatstyle{ruled}
\newfloat{listing}{tb}{lst}{}
\floatname{listing}{Listing}
\usepackage{subfig}
\usepackage[most]{tcolorbox}
\tcbset{colframe=white, boxrule=0pt}
\usepackage{amssymb}

\title{The Dilemma Between Data Transformations and\\Adversarial Robustness for Time Series Application Systems}
\author{
    Sheila Alemany,
    Niki Pissinou
}
\affiliations{
    School of Computing and Information Sciences,\\
    Florida International University\\
    salem010@fiu.edu, pissinou@fiu.edu
}

\begin{document}

\maketitle

\begin{abstract}

Adversarial examples, or nearly indistinguishable inputs created by an attacker, significantly reduce machine learning accuracy. Theoretical evidence has shown that the high intrinsic dimensionality of datasets facilitates an adversary's ability to develop effective adversarial examples in classification models. 
Adjacently, the presentation of data to a learning model impacts its performance. For example, we have seen this through dimensionality reduction techniques used to aid with the generalization of features in machine learning applications. Thus, data transformation techniques go hand-in-hand with state-of-the-art learning models in decision-making applications such as intelligent medical or military systems.
With this work, we explore how data transformations techniques such as feature selection, dimensionality reduction, or trend extraction techniques may impact an adversary's ability to create effective adversarial samples on a recurrent neural network. Specifically, we analyze it from the perspective of the data manifold and the presentation of its intrinsic features.
Our evaluation empirically shows that feature selection and trend extraction techniques may increase the RNN's vulnerability. A data transformation technique reduces the vulnerability to adversarial examples only if it approximates the dataset's intrinsic dimension, minimizes codimension, and maintains higher manifold coverage.

\end{abstract}

\section{Introduction}
\label{sec:introduction}


As the application of ML grows in industries that require explainable and reliable ML models, there is a significant concern on the immense fragility in neural networks when given a varying size set of imperceptibly perturbed inputs, adversarial examples \cite{biggio2018wild, su2019one, elsayed2018adversarial}.
To address this issue, many pioneering works have focused on solutions that increase the models' robustness to maintain high accuracy assuming the existence of these adversarial examples \cite{biggio2018wild, ilyas2019adversarial, goodfellow2018making, hendrycks2021unsolved}. 
The solutions proposed in these works have observed adversarial examples from the perspective of the abstractions created by the machine learning models. But, since these datasets are incomplete and instantaneous representations of information, trained machine learning models contain many areas within it with low confidence. These low confidence areas of knowledge can be mapped similarly to how a human can be less sure of a correct answer for unfamiliar contexts. Adversaries exploit these low-confidence areas and create a minor input change possible to skew the model's recommendations or decisions to be wrong or inaccurate. Despite these observations \cite{ilyas2019adversarial, goodfellow2018making}, the existence of adversarial examples remains an open problem \cite{shafahi2018adversarial, hendrycks2021unsolved}. However, these proposed theories continuously approach similar conclusions: \textit{the vulnerability of ML models is highly correlated to how the data is represented}.

In practice, data is repeatedly being transformed with a growing list of pre-processing techniques to optimize ML models \cite{aleman2018using, naranjo2019fuzzy, huang2019re}, and these techniques transform the way data is presented to an intelligent system. 
Thus, based on existing work, we hypothesize that data transformations may directly impact the adversary's ability to create adversarial samples due to manipulations in representing the intrinsic features of data. Motivated by the direct impact that this may have on currently deployed systems, we explore how five widely-applied data transformation techniques affect the robustness\footnote{In this work, robustness refers to the adversary's decreased capacity to attack more efficiently or induce inaccurate results using "harder-to-detect" perturbations.} of recurrent neural networks. 

We consider techniques that span three different data transformation categories: \textit{dimensionality reduction} (Principal component analysis \cite{shlens2014tutorial}), \textit{feature selection} (random forest \cite{golay2017unsupervised} and low variance \cite{bramer2004artificial}), and \textit{trend extraction} (candlestick charting \cite{chmielewski2015pattern} and exponential moving average \cite{klinker2011exponential}).
Our empirical evaluation aims to identify whether data transformation techniques in the three categories \textit{can} impact the efficiency of an adversarial attack. To better understand this, we design our experiments to explore the following questions: 
\begin{enumerate}
    \item Could data transformations contribute to any adversary's ability to more easily construct adversarial examples (i.e., make the ML model more vulnerable to attacks)?
    
    \item Is the dimensionality reduction technique, PCA, consistent as a strategy to increase robustness, as seen in \citet{bhagoji2018enhancing}, when given a time series dataset, recurrent neural network, and varying selected principal components?
    
    \item What representations of data contribute to ML models that are least susceptible to adversarial examples and how can we use them to ensure best practices when manipulating data?
\end{enumerate}
Overall, in this work, we expand the empirical understanding of how data transformation techniques may impact the robustness of a recurrent neural network given the Carlini \& Wagner \cite{carlini2017towards} evasion attack on a multi-variate time series dataset \cite{banos2015mdurance}. This benefits ML practitioners as they can use the presented results to move towards better data practices when manipulating data increasingly used in deployed intelligent systems. This is the first work exploring whether certain data transformations (outside of dimensionality reduction) may impact robustness in time series ML models to the best of our knowledge.


\section{Related Work}
\label{sec:related-work}


Many pioneering works have established a foundation for the seemingly inherent vulnerability to adversarial examples. 
\citet{szegedy2013intriguing} argued the existence of low-probability adversarial ``pockets'' that an adversary can take advantage of. 
\citet{feinman2017detecting} established that adversarial samples lie furthest away from the data manifold\footnote{Data manifold is defined as the geometry of the data which contains a topological space that locally resembles the Euclidean space near each data value.} and are restricted in the direction normal to the data manifold such that the adversarial examples cross the decision axis (the optimal boundary between the data manifolds captured during model training time) and result in an incorrect output \cite{khoury2018geometry}. 

\citet{shafahi2018adversarial} and \citet{ilyas2019adversarial} proposed that the vulnerabilities to adversarial examples stem from the foundational characteristic in ML that the training data accurately and adequately represents the underlying and \textit{abstracted} phenomena through the learning process. Such high dimensional abstractions\footnote{Highly dimensionality of a model is not only correlated to the model architecture/parameters but also the dataset being used \cite{su2019one}.} allow adversaries to exploit through minor and specific details that a trained ML model can overlook. 
Similarly, \citet{amsaleg2020high} showed that the intrinsic dimensionality of datasets and an adversary's ability to develop effective adversarial examples are directly proportional in classification models. This is so as a higher intrinsic dimensionality results in higher model complexity. 
In all cases, the quality of the abstractions is limited to how the data is presented to the model (i.e., does the data have bias? Is it missing values? Does it contain noise? etc.). This is because ML learning/generalization and adversarial example creation remains a classic optimization problem.

Data dimensionality has been referred to as a ``curse'' due to substantial computational complexity yielding difficulties when abstracting properties in data that do not occur in lower-dimensional data \cite{van2009dimensionality, ilyas2019adversarial, bhagoji2018enhancing}. Resulting in data transformations techniques often being used in learning systems to improve upon these burdens \cite{cheng2018novel}. 
Naturally, data transformations have influenced the field of adversarial ML due to the connection between adversarial vulnerability in deep learning and the high dimensionality of data.
These techniques increase robustness by modifying the input such that the impact of gradient-based attacks is reduced, either through adversarial pre-training \cite{hendrycks2019pretraining}, feature squeezing \cite{xu2017feature}, dimensionality reduction with PCA \cite{bhagoji2018enhancing}, or identifying and removing the least ``robust features'' which contribute the most to a model's vulnerability \cite{ilyas2019adversarial}. Thus, they are defenses that focus on executing certain transformations at the beginning of the ML pipeline, such that when the adversary gains perfect knowledge of the trained model, it is more difficult for an adversary to optimize its attack.


\citet{carlini2017adversarial} showed how certain previously described techniques, including \cite{bhagoji2018enhancing}, were not a consistent defense. For example, they were able to show how using PCA in the training data did not increase the robustness of a convolutional neural network, only the fully-connected network. Other works had inconsistencies in their presented results when tested on other datasets.
Observing these inconsistencies and how the representation of data highly influences abstractions, we hypothesize that different data transformations may individually impact the representation of the intrinsic features and hence, uniquely impact an adversary's ability to attack the model.

\section{Data Transformation Techniques}
\label{sec:data-transformations}

Our comparative review includes data transformation techniques during the pre-processing stage of the ML pipeline. 
It is \textit{not} exhaustive. We have strictly focused on linear data transformation techniques that have been commonly used in a variety of applications \citep{aleman2018using, bhagoji2018enhancing, carlini2017adversarial}. For brevity, we assume the reader understands the each technique. Future work can be focused on non-linear dimensionality reduction techniques. We keep both works separate as non-linear transformations may impact the complexity of data manifolds differently than linear ones.


\paragraph{Dimensionality Reduction}
\label{sec:pca}

Dimensionality reduction is the transformation of high-dimensional data into a significant representation of low dimensionality \citep{cheng2018novel}. Principal component analysis (PCA) is by far one of the more popular unsupervised tools due to its simple, non-parametric method for extracting relevant information from overwhelming datasets \citep{shlens2014tutorial}. 
For this work, we consider using 27\%, 50\%, and 81\% of the principal components to approximate the feature counts around the 25, 50, and 75 quartiles. We explore in Section \ref{sec:evaluation-results} how the selected principal components in varying extremes can significantly change the data manifold in ways which impact robustness.


\paragraph{Feature Selection} 

Feature selection is a data transformation technique that has been used for decades to represent particular relationships in data by eliminating features that may be irrelevant or redundant \citep{dash1997feature} based on a varying set size of heuristics. These techniques compare to dimensionality reductions methods in that they do not map onto a lower-dimensional space. 
For this work, we have selected random forest selection \citep{golay2017unsupervised} and low variance selection \citep{bramer2004artificial} due to their high usage for their low computational requirements. 

For random forest selection, we set the feature importance measure threshold to be the mean of all importance values, as it is standard in practice \citep{golay2017unsupervised}.
For low variance selection, the selected features contributed 91.1\% of the total variance in the data, as it is said to be the best heuristic to approximate the most significant information of a dataset \citep{van2009dimensionality}. 
Although random forest selection considers the relationship of features with the target variable and low variance selection does not, both techniques chose 9 overlapping features. Thus, we expect their impact on data manifolds to be similar even with their varying heuristics for feature selection.


\paragraph{Trend Extraction}
\label{sec:timeseries}

Up-to-date works have focused on image recognition tasks concerning robustness, but time series data is also highly used in ML applications. As a result, we have analyzed the impact of data transformation techniques meant to extract trends in time series data, such as candlestick charting \cite{chmielewski2015pattern} and exponential moving average (EMA) \cite{klinker2011exponential}. 

These techniques were selected as they are used in prediction tasks in areas such as financial markets \cite{naranjo2019fuzzy}, IoT \cite{aleman2018using}, and object tracking \cite{huang2019re}. These techniques affect the data manifold by smoothing the trends in time series data, similarly to feature squeezing for image recognition \cite{xu2017feature}, by artificially reducing the distance between temporally adjacent points that provide better estimation of their distance along the manifold.
For this work, to ensure we are similarly comparing both trend extraction techniques, both were assigned the same value for the time window. The time window value of 20 was a selected hyperparameter that would not reduce the dimensionality of the dataset enough to hinder the model accuracy for the candlestick charting technique but would cause a significant enough change to the feature trends given the EMA technique.


\section{Threat Model}
\label{sec:threat-model}

As per \citet{carlini2019evaluating}, we define the adversary's knowledge, capabilities, and goals to ensure analysis for worst-case robustness. We did not implement any additional defenses as our goal for this work is to explore the impact of these techniques for small perturbation budgets that are difficult to detect using the current state-of-the-art defenses \cite{tjeng2017evaluating}. Considering the attack success rate with incorporated defenses and data transformation techniques is left for future work.

\paragraph{Knowledge} We use a white-box attack where the adversary has full access to the trained neural network model, the defense used, along with the data distribution at test time. We consider this attack because white-box attacks are more powerful than black-box attacks, as a white-box attack can reach a 100\% success rate. Additionally, we consider evasion attacks where the adversaries can attack only during model deployment, meaning that they tamper with the input data after the deep learning model is trained. 



\paragraph{Capabilities} For the attack method, we use the iterative optimization-based method of \citet{carlini2017towards}. 
We selected this attack model due to its high success at crafting effective adversarial samples with the lowest distortion \cite{carlini2017towards}. Specifically, we have used the Carlini \& Wagner $l_{\infty}$ implementation from the Adversarial Robustness Toolbox by IBM Research \cite{art2018}. Some minor hyperparameters were modified to create adversarial attacks that reduced the accuracy of our model are the learning rate and confidence, set to 0.01 and 0.5, respectively. 


\paragraph{Goal} To create effective adversarial examples, we use the $l_\infty$ distortion metric to measure the similarity between the benign and potential adversarial examples since the $l_\infty$-ball around each data point has recently been studied as an optimal, natural notion for adversarial perturbations \citep{goodfellow2014explaining, carlini2017towards}. For this work, we used the untargeted attack and considered $0 < \epsilon \leq 1$ \citep{tjeng2017evaluating}. Although targeted attacks are more powerful concerning the attack success rate, we are considering an untargeted attack since these attacks require a more limited perturbation budget that allows for an adversary to efficiently deploy the attack undetected \citep{carlini2017towards}. 
We can visualize the perturbation under this distance metric by viewing a series of data points. There is a maximum perturbation budget of $\epsilon$, where the sum of all perturbations is allowed to be changed by up to $\epsilon$, with no limit on the number of modified values. Since perturbation budget has to remain less than some small $\epsilon$, even if all values are modified, the trends in time series data will appear visually identical.


\begin{figure*}[h]
\centering
\subfloat[MHealth Dataset]{%
\includegraphics[width=0.25\linewidth]{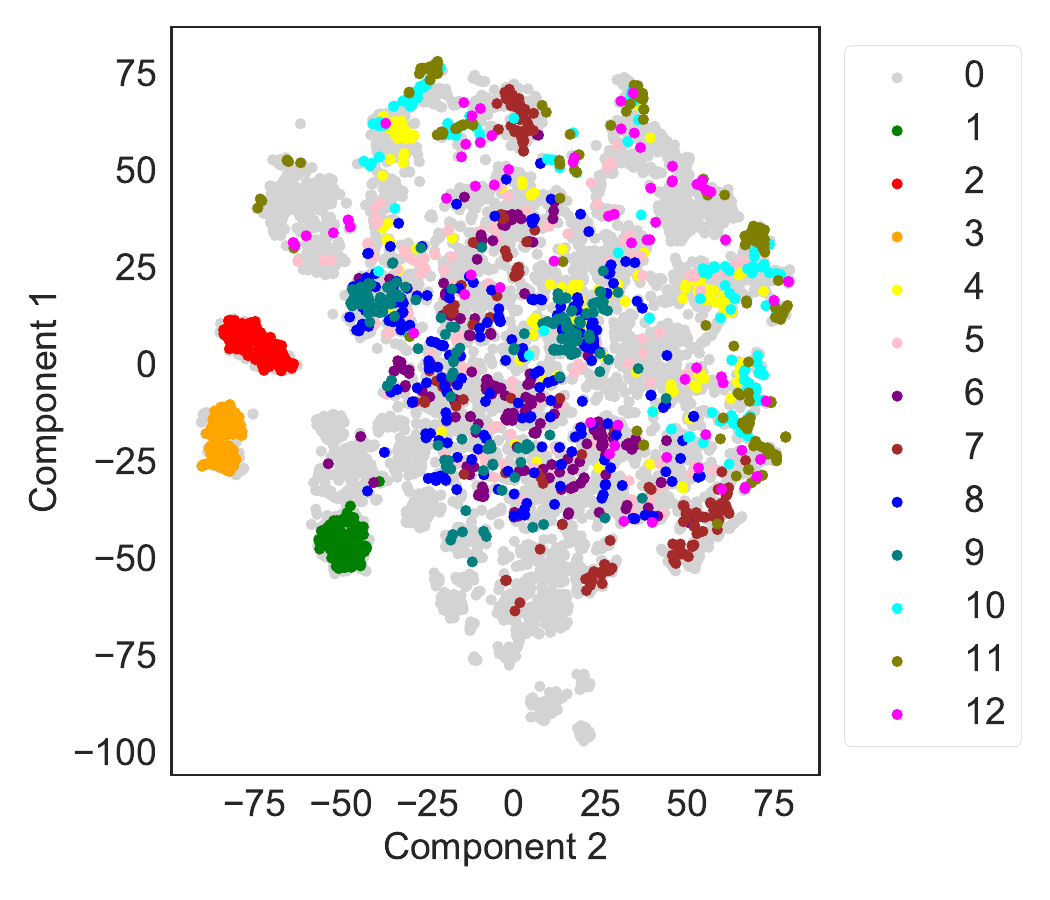}
  \label{fig:mhealth1}%
}
\subfloat[MNIST Dataset]{%
\includegraphics[width=0.25\linewidth]{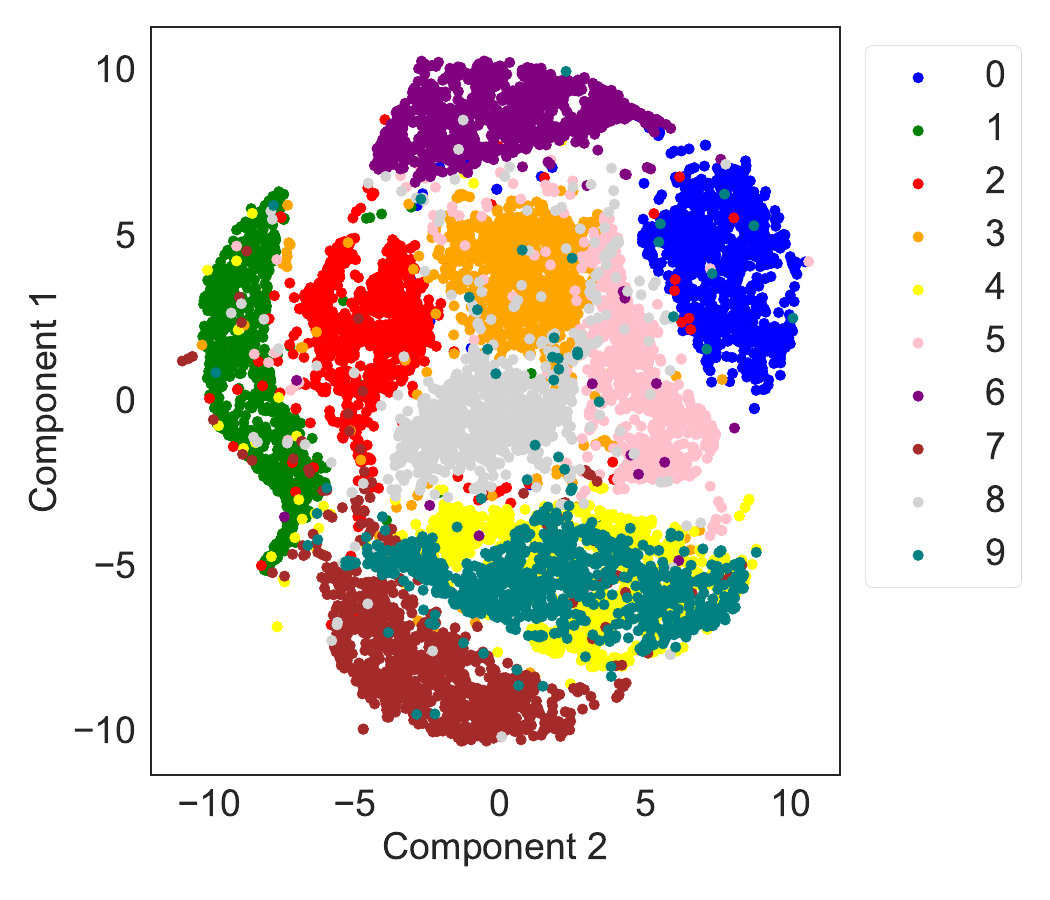}
  \label{fig:mnist1}%
}
\subfloat[CIFAR-10 Dataset]{%
\includegraphics[width=0.25\linewidth]{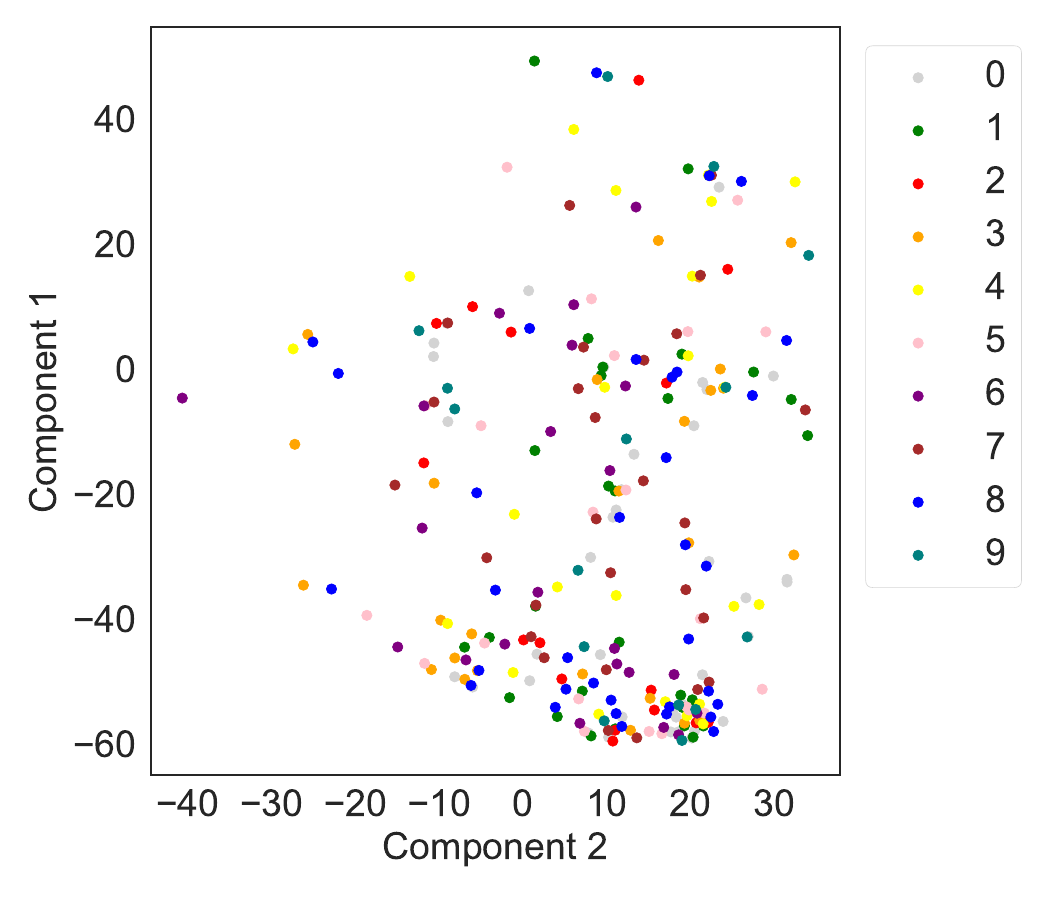}
  \label{fig:ciphar1}%
}
\caption{Visualization of datasets using T-SNE to observe the relationships between the points in high-dimensional space using 1000 randomly selected points from each dataset. MHealth shows that various clusters can be easily identified, such as the points in classes 1, 2, and 3, similar to MNIST. Yet, there are clusters such as for classes 8 and 12, where the points are more scattered, similar to CIFAR-10.}
\label{fig:tsne}
\end{figure*}



\section{Experimental Methods}
\label{sec:experimental-methods}

We compare our evaluation results with previous works that have completed similar tests with the computer vision datasets, CIFAR-10 \cite{Krizhevsky09learningmultiple} and MNIST \cite{lecun2010mnist}, to check for overall consistency in the impact done by data transformation techniques. 

\paragraph{Dataset}
\label{sec:dataset-details}

The focus of related adversarial evaluation is largely centered around image recognition tasks. However, there are high dimensional time series datasets that have received little attention in the adversarial ML field and the need for evaluation on other datasets is crucial for the advancement of the area \citep{carlini2017adversarial}. As a result, we have used the MHealth (Mobile Health) Dataset\footnote{Dataset available on the UCI ML Repository at \url{https://archive.ics.uci.edu/ml/datasets/MHEALTH+Dataset}} which contains body motion and vital signs recording of individuals while performing several physical activities \citep{banos2015mdurance}. This highly volatile dataset contains 22 total features which map to one of 12 potential physical activities and we selected the data corresponding to \texttt{subject1} with a total of 160,860 timestamps.

Figure \ref{fig:mnist1} shows that the MNIST dataset contains the most well-defined classes meaning points corresponding to the same class are clustered together more frequently. This implies that the points within each class of the MNIST dataset have highly correlated relationships even with the highly-dimensional dataset. On the contrary, in Figure \ref{fig:ciphar1}, the CIFAR-10 dataset does not have well-defined clusters resulting in an almost opposite conclusion relative to the MNIST dataset. As a result, CIFAR-10 has been described as a substantially more difficult dataset to work with. Therefore, conclusions made with MNIST may contain properties that do not generalize across tougher datasets such as CIFAR-10 \citep{carlini2017adversarial}. However, the MHealth dataset lies between the MNIST and CIFAR-10 dataset in regards to the relationship between the points in high-dimensional space. 
Thus, we are testing with a realistic time series dataset that contains manifold properties that may carry-out to various other highly-dimensional time series datasets. As a result, we believe our evaluation using the MHealth dataset is a valid example that brings to light the observations presented in this work. 

\paragraph{Learning Model}
\label{sec:model-details}

Data pre-processing includes processes such as data cleaning, normalization, transformation, feature extraction, selection, and is the step done before training in this work. 
For the learning model, we have implemented a multi-class classification recurrent neural network (RNN) with LSTM layers using Keras \cite{chollet2015keras}. Network architecture and hyperparameter tuning were completed to guarantee that all trained models for each data transformation technique received the same hyperparameters while maintaining testing accuracy above 90\% to ensure that the network architecture did not influence robustness results. The network contained contain only two LSTM units combined with dropout layers which showed to return satisfactory training and testing results.
We used the hyperbolic tangent function in these hidden vectors as it is a standard activation function among recurrent neural networks \citep{chollet2015keras}. The dropout values were set to 0.1, depicting that 10\% of each input was ignored to prevent the model from overfitting to the training data. Lastly, we are not concerned about our network's simple linear structure because it is claimed that the network's simple structure architecture does not impact their Carlini \& Wagner evasion attacks \citet{carlini2017towards}. 

\begin{figure*}[t]
    \centering
    \includegraphics[width=0.78\linewidth]{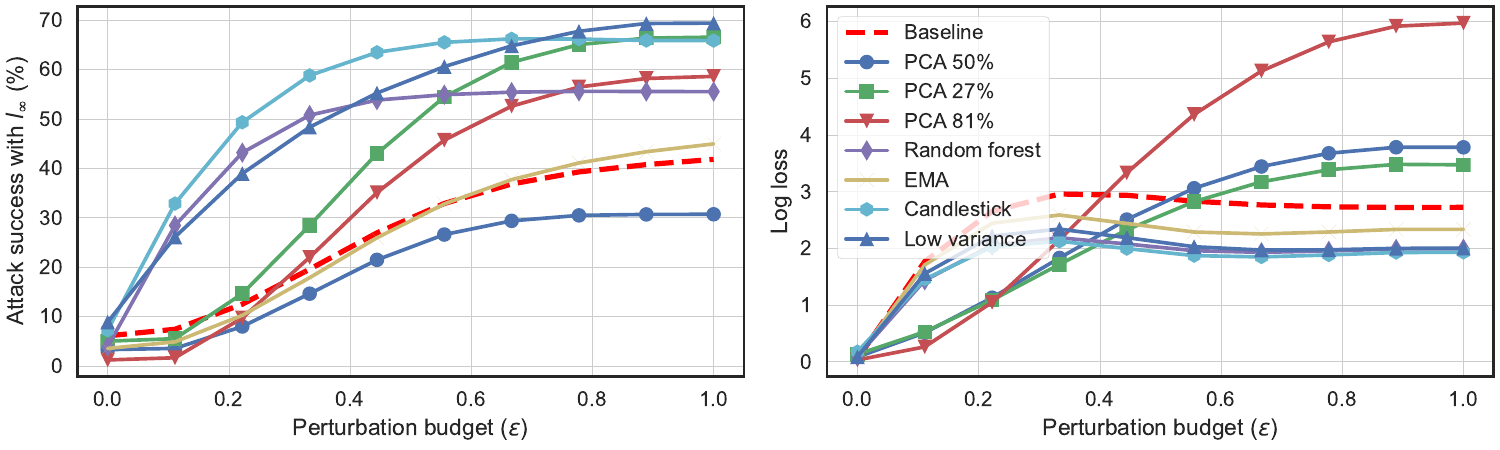}
    \caption{Attack success and log loss scores given five data transformation techniques against the baseline model without pre-processing. We can see that the best performing technique was PCA using half of the principal components. However, the log loss scores corresponding to model confident shows the all PCA techniques returned the lowest confidence when $\epsilon > 0.57$.}
    \label{fig:attack-logloss}
\end{figure*}

\begin{figure*}[t]
    \centering
    \includegraphics[width=0.78\linewidth]{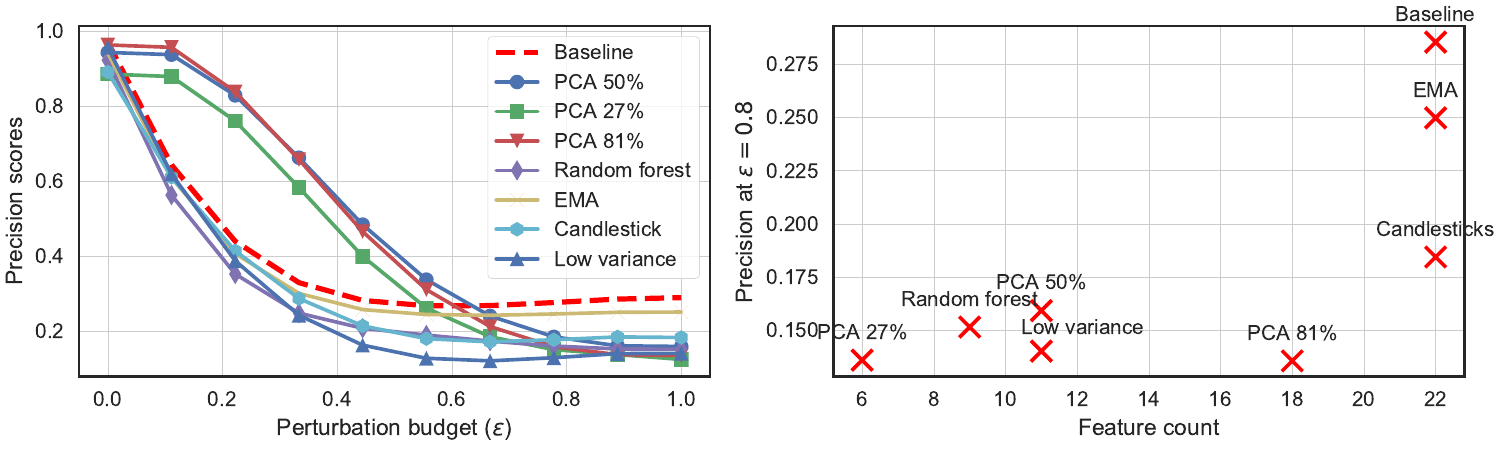}
    \caption{Precision scores under-performed for all techniques once the perturbation budget was over $\epsilon = 0.68$. From the scatter plot, we can see that reducing the number of features during training negatively impacted the precision scores given a high enough perturbation budget.}
    \label{fig:precision}
\end{figure*}



\section{Robustness Against Evasion Attacks}
\label{sec:evaluation-results}

Since the data manifold structure heavily influences the existence of adversarial examples and how these adversarial attacks are optimized, we observe the changes in model performance from the perspective of the data manifold. To compare the changes made to the manifold by the data transformations, we observe the codimension or the difference between the dimension of the data manifold and the dimension of the embedding space \footnote{The embedding space is the space in which the data is embedded after dimensionality reduction.} \citep{khoury2018geometry}. We show only perturbation budget 0 to 1 to show the impact given small perturbations since the concluding results do not change as the attack success continues increasing.



\begin{table*}
  \caption{Summary of results: Columns from left to right present the data transformation technique, the number of features used from the original data, its clean accuracy when the model is not under attack, the perturbation budget required to attack success to 30\%, and percentage change in robustness at $\epsilon = 0.80$ relative to the baseline model with no data transformation applied to its training data.}
  \label{sample-table}
  \centering
  \begin{tabular}{ccccc}
    \hline
    Data Transformation & Feature Count & Benign Accuracy   & Distance ($l_{\infty})$  & $\Delta$ in Robustness \\
    \hline
    Baseline    & 22    & 97.93\%   & 0.51  & - \\
    PCA 50\%        & 11    & 96.71\%   & 0.40  & $\uparrow$ 24.39\% \\
    PCA 81\%        & 18    & 98.80\%   & 0.76  & $\downarrow$ 43.90\% \\
    PCA 27\%        & 6     & 95.00\%   & 0.34  & $\downarrow$ 60.98\% \\
    Random Forest & 9   & 96.11\%   & 0.13  & $\downarrow$ 31.71\% \\
    Low Variance & 11   & 91.32\%   & 0.15  & $\downarrow$ 65.85\% \\
    Candlesticks & 22   & 92.78\%   & 0.11  & $\downarrow$ 60.98\% \\
    EMA         & 22    & 96.48\%   & 0.51  & $\downarrow$ 7.32\% \\
    \hline
  \end{tabular}
\end{table*}

\subsection{Manifold Impacts on Log Loss \& Precision} 

From Figure \ref{fig:precision} on the next page, we can see that precision is consistently below baseline for both feature selection and trend extraction techniques. The low log loss and precision indicates that these models are overly confident but erroneous implying a closer proximity between submanifolds to the decision axis \citep{wu2017manifold}. In other words, when the submanifolds are closer to the decision boundary, the distance between two arbitrary points in different classes is lower relatively. Thus, when an ML model is tasked with categorizing a new point, it will often confidently miscategorize since it is ``harder" to differentiate between the two candidate classes. From the perspective of an adversary, they now require a minimal perturbation budget to ``convince" the ML model to miscategorize incoming data points consistently with high confidence. However, this is not the case with PCA. With PCA, the precision is improved when $\epsilon < 0.65$ due to relatively better defined submanifolds as a direct result of mapping the input embedding into a lower dimension. The reduced precision for greater values of epsilon is then introduced when the log loss of the model increases because linear units can get low precision from responding too strongly from a reduced confidence when it does not understand samples with larger perturbations \citep{goodfellow2014explaining}.

\begin{tcolorbox}
\textit{Takeaway 1.1: PCA creates more well-defined submanifolds for each class such that it is more difficult for an adversary to ``trick" an ML model with an imperceptible adversarial example. This is not the case for feature selection and trend extraction techniques.}
\end{tcolorbox}

\subsection{Manifold Impacts on Model Accuracy} 

From Figure \ref{fig:attack-logloss}, it is clear the attack success rate is only hindered by 24.39\% when the PCA technique is used with half of its principal components. \citet{bhagoji2018enhancing} proposed that PCA should consistently increase robustness because PCA removing the high variance components should eliminate the features that adversaries can easily take advantage of. However, as \citet{carlini2017adversarial} already showed this is not consistent given a convolutional neural network and, for our evaluation, it seems it is may not always consistent with our recurrent neural network.

The other PCA techniques using 27\% and 81\% of the principal components did not perform as well once the perturbation budget exceeded $\epsilon = 0.1$. Particularly, using only 27\% of the principal components results losing too many dimensions which can in turn reduce the manifold coverage for the dataset. This lack of coverage makes it is much easier for an adversary to find an example far away from the data manifold \cite{feinman2017detecting}. This can happen easily in practice since high training/testing accuracy does not imply high accuracy/coverage of the data manifold \cite{khoury2018geometry}. On the other hand, when using 81\% of the principal components, there is high codimension resulting in relatively more directions normal to the manifold and directly contributing to a more efficient attack. Thus, we can conclude that an optimal codimension exists in datasets such that the vulnerabilities presented are minimized. 

\begin{tcolorbox}
\textit{Takeaway 2.1: The dimensionality reduction technique, PCA, is not a consistent defense against adversarial examples when the codimension is not optimal.}
\end{tcolorbox}

The feature selection techniques behaved similarly (as expected) given both techniques selected a majority of the same features. 
In both cases, since no mapping to a lower dimension occurs and a majority of the features are removed, the model contains high codimension and a lack of manifold coverage relative to the dimensionality reduction. As a result, feature selection aids an efficient adversarial attack through all tested perturbation budgets.

\begin{tcolorbox}
\textit{Takeaway 2.2: Feature selection techniques contribute to higher codimension, and they lack manifold coverage results in an adversary's ability to construct adversarial examples more easily.}
\end{tcolorbox}

The trend extraction techniques, however, do not remove the features used but manage to force the data into a lower dimensional manifold by generalizing the trends that normally contribute to the high dimensionality in trained models \citep{xu2017feature}. 
For the candlestick charting, the transformation into the four-tuple reshaped the features but contributed to fundamental information loss for the dataset. The information loss resulted on the higher relative end of codimension and the one of most efficient creation of adversarial examples with a 60.98\% decrease in robustness at $\epsilon = 1.0$.
However, EMA seemed to not smooth the manifold enough for a drastic change from the baseline data. Therefore, no statistically significant change to the data manifold results in a performance on par with the baseline.

\begin{tcolorbox}
\textit{Takeaway 2.3: Candlesticks charting contributes to the most vulnerable ML models due to information loss which significantly increases codimension.}
\end{tcolorbox}


\subsection{Optimal Data Representations}
\label{sec:intrinsic}


From our experimentation, we were able to see that the data transformation techniques which did not minimize codimension aided in allowing pathways for adversaries to exploit. The difficulty arises when transformations do not always and consistently impact the codimension. This prompted us to ask the following question: \textit{how do we know how and what transformation to execute to ensure that the codimension is not increased for an arbitrary dataset?}

Reaching this ideal data representation can be done by identifying the intrinsic dimension of a dataset. The intrinsic dimension is defined as a potential solution from the codimension of solutions sets \citep{li2018measuring}. In other words, it can be described as the minimum number of parameters necessary to account for the observed properties in the data, achieve optimal ML performance accuracy and, a way to reduce codimension.

\begin{tcolorbox}
\textit{Takeaway 3.1: ML practitioners can reduce codimension in their models using the intrinsic dimension of their dataset.}
\end{tcolorbox}

\subsubsection{Finding and Using Intrinsic Dimension}
\label{sec:finding-intrinsic}

The geometry of the data manifold, or the dataset's intrinsic dimensionality, is generally twisted and curved with non-uniformly distributed points, making identifying the intrinsic dimensionality a challenging task unique for each dataset \citep{facco2017estimating}. There are various tools and algorithms to analyze the intrinsic characteristics, such as the intrinsic dimensionality of data. 
For example, the most straightforward way by counting the number of features that contribute at least 90\% of the total variance \cite{van2009dimensionality}. 
With datasets and ML models that are more complex, \citep{li2018measuring} proposed to measure the intrinsic dimension of an ``objective landscape" or the dimension of the subspace of a parameterized model, such as a dataset or neural network. They do so by training a neural network from a small, randomly oriented subspace and slowly increasing its dimension (through added features or parameters) until they reach a plateau of performance accuracy, and define that configuration to be the objective landscape's intrinsic dimension.

To measure the intrinsic dimension of the MHealth dataset, we used both of these techniques, \citep{van2009dimensionality} and \citep{li2018measuring}. Using \citep{van2009dimensionality}, 11 features contribute to approximately 91\% of the total variance. Using \citep{li2018measuring}, we sorted the features by descending variance and trained the same RNN one feature at a time and noticed the plataeu began with 9 features at approximately 94\% performance accuracy. Overall, from these simple tests, we can see that the intrinsic dimension for the MHealth dataset is approximately between $[9, 11]$, likely closer to 9 due to the complexity of the model and the looser bounds presented by the \citep{van2009dimensionality} heuristic. 
Since the same neural network architecture and parameters are used for all transformation techniques, its contribution to the intrinsic dimensionality is out of scope of this evaluation. However, ML practitioners can incorporate the technique for future parameter configurations into their pipelines with ease.

\begin{tcolorbox}
\textit{Takeaway 3.2: Observing the objective landscape of data is one simple, flexible, and accurate way to identify the intrinsic dimension for consideration along with any data transformations.}
\end{tcolorbox}

\subsubsection{Intrinsic Dimension on Robustness with MHealth}

With the dimensionality reduction technique, PCA, we were able to see that the performance was only consistent in the case when the input embedding dimensionality more closely approached the intrinsic dimension. Given the intrinsic dimensionality reached with PCA 50\%, the codimension was relatively minimized resulting in the most restricted number of directions for the adversary to take advantage. 

On the other hand, for the feature selection techniques, the lack of mapping to a lower dimension prevented the feature selection techniques to approximate the intrinsic dimension as accurately as PCA, resulting in the poor performance while under attack. However, since random forest selection closer approximates the intrinsic dimension (with 9 selected features), the attack success rate differs to low variance selection by approximately 10\%. Also, for the candlesticks, the transformation into the four-tuple strayed the furthest away from the intrinsic dimensionality by reshaping the features. This transformation contributed in fundamental information loss for the dataset while straying away from the intrinsic dimension resulting on the higher relative end of codimension and one of the most efficient creation of adversarial examples with a 60.98\% decrease in robustness at $\epsilon = 1.0$. 

\begin{tcolorbox}
\textit{Takeaway 3.3: To avoid introducing additional vulnerabilities in ML pipelines, one must observe and understand the particular dataset's intrinsic characteristics and ensure any transformation does not stray from the intrinsic dimension.}
\end{tcolorbox}

\section{Conclusion}

For this work, we have provided an example where linear data transformation techniques can change an adversary's ability to create effective adversarial examples. From the conclusions presented in \citet{amsaleg2020high}, one could be led to believe a transformation that has reduced complexity and high training/testing accuracy would be inherently more robust. However, their conclusion stands between datasets of different complexities but does not speak on the potential impacts of data transformations. Positive impacts by dimensionality reduction techniques are only presented where the technique embeds the high-dimensional input space into a lower-dimensional structure \textit{that approaches the intrinsic dimension of data}. Specifically, PCA overperformed \textit{only} when the dimensionality approached the intrinsic dimension. Meanwhile, the trend extraction techniques that refrained from sufficiently reaching the intrinsic dimension showed to negatively impact the attack success and the precision scores, overall making the ML model more vulnerable to adversarial examples. Although we only considered a recurrent neural network with LSTM layers, the MHeath dataset that we used is a realistic, high-dimensional time series dataset that shows an example of the impacts that data transformation can have on an ML model.

Our results conclude that when the dimension approaches the optimal intrinsic dimension, lower codimension and higher manifold coverage result in a lesser need to generalize features and reduce the inherent vulnerability to adversarial examples.
However, it is important to note that reaching the intrinsic dimensionality is not enough to guarantee perfect robustness. The inevitability of adversarial examples has recently been theoretically studied, and it is still not possible to know the exact and consistent properties of real-world datasets or the resulting fundamental limits of adversarial training for specific datasets \citep{shafahi2018adversarial}.
In other words, the underlying distributions themselves can be complex enough such that there may be no guarantee of perfect robustness against adversarial examples. Nonetheless, our work highlights the value of considering potential vulnerabilities introduced to ML pipelines through data transformations and how ML practitioners may utilize the intrinsic dimension to reduce the overall complexity of models, avoid introducing additional vulnerabilities, and create more reliable pipelines.

Lastly, as a future direction, the analysis of data transformations (linear and non-linear) on adversarial examples may benefit a model under a poisoning attack. Such analysis could provide insight into how certain data transformations can extricate adversarial noise to increase model robustness.

\bibliography{aaai22}

\end{document}